\documentclass[letterpaper, 10 pt, conference]{ieeeconf}  
\IEEEoverridecommandlockouts                              
\usepackage{epsfig} 
\usepackage[left=19mm, right=19mm, top=19mm, bottom=19mm]{geometry}
\usepackage{times}  
\usepackage{url}  
\usepackage{graphicx}  
\usepackage{times}
\usepackage{url}
\usepackage{graphicx} 
\usepackage{listings}
\usepackage{amsmath,amssymb,xspace,microtype}
\usepackage[pagebackref=true,breaklinks=true,colorlinks,bookmarks=false]{hyperref}
\DeclareMathOperator{\E}{\mathbb{E}}

\title{\LARGE \bf
Unsupervised Representation Adversarial Learning Network: from Reconstruction to Generation
}

\author{Yuqian Zhou*\thanks{*Authors have equal contribution}, Kuangxiao Gu*
, Thomas Huang\\
IFP, Beckman Institute, UIUC\\
$\{$yuqian2, kgu3$\}$@illinois.edu, huang@ifp.uiuc.edu\\
}

\begin{document}

\maketitle
\thispagestyle{empty}
\pagestyle{empty}

\begin{abstract}
A good representation for arbitrarily complicated data should have the capability of semantic generation, clustering and reconstruction. Previous research has already achieved impressive performance on either one. This paper aims at learning a disentangled representation effective for all of them in an unsupervised way. To achieve all the three tasks together, we learn the forward and inverse mapping between data and representation on the basis of a symmetric adversarial process. In theory, we minimize the upper bound of the two conditional entropy loss between the latent variables and the observations together to achieve the cycle consistency. The newly proposed RepGAN is tested on MNIST, fashionMNIST, CelebA, and SVHN datasets to perform unsupervised classification, generation and reconstruction tasks. The result demonstrates that RepGAN is able to learn a useful and competitive representation. To the author's knowledge, our work is the first one to achieve both a high unsupervised classification accuracy and low reconstruction error on MNIST. Codes are available at \href{https://github.com/yzhouas/RepGAN-tensorflow}{https://github.com/yzhouas/RepGAN-tensorflow}.

\end{abstract}

\section{Introduction}
Learning a good representation from complex data distribution can be resolved by deep directed generative models. Among them, Generative Adversarial Network (GAN)\cite{goodfellow2016nips} is proposed to generate complicated data space by sampling from a simple pre-defined latent space. Specifically, a generator is modeled to map the latent samples to real data, and a discriminator is applied to differentiate real samples from generated ones. However, the original GAN only learns the forward mapping from a entangled latent space to data space. Given the complicated data, it lacks the inverse inference network to map the data back to the interpretable latent space. 

\begin{figure}[t]
	\begin{center}
		\includegraphics[width=1\linewidth]{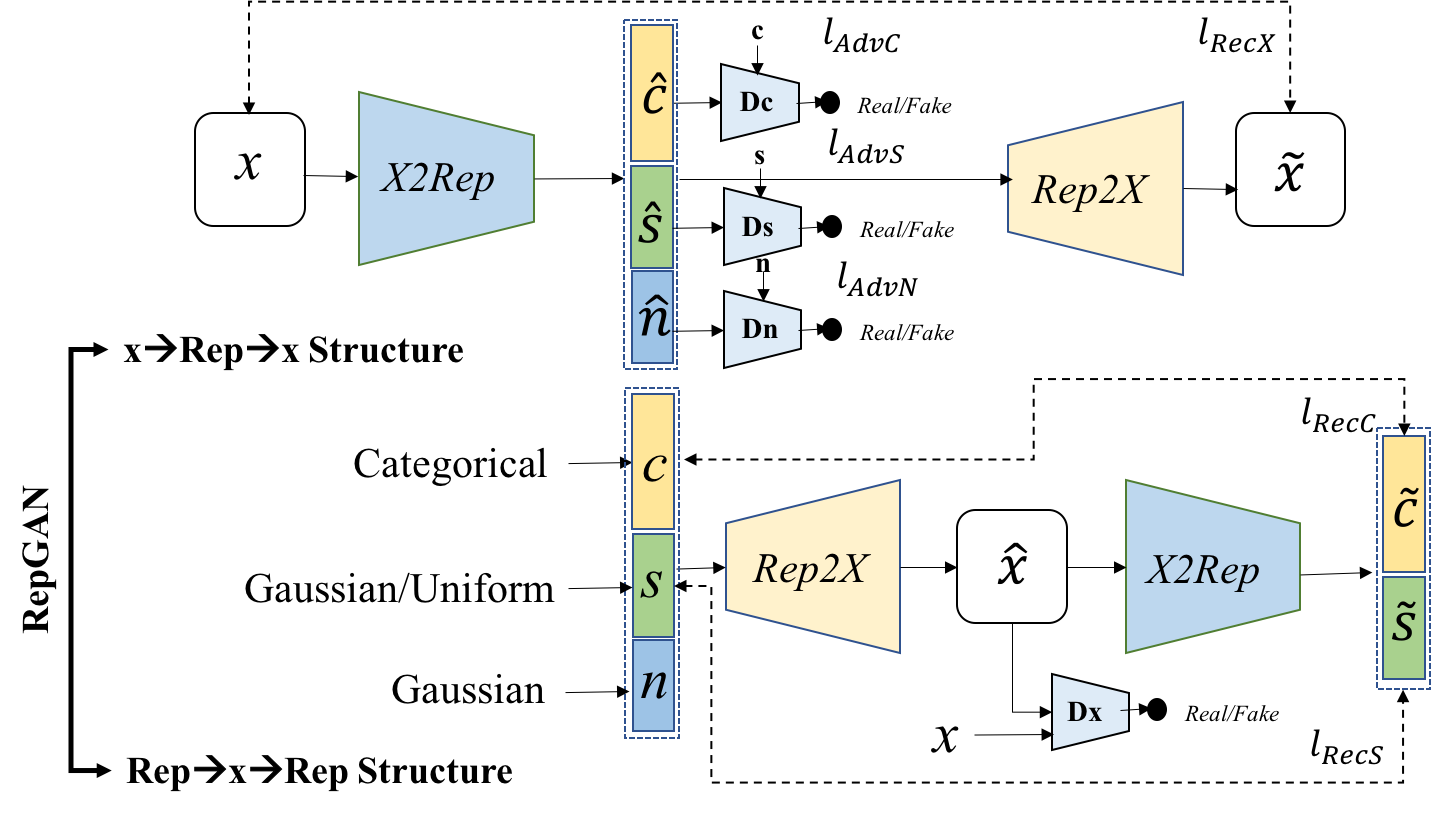}
	\end{center}
	\caption{Structure of the proposed Representation GAN (RepGAN). The latent representation is split to a categorical slot $c$, continuous slot $s$, and noise $n$. RepGAN jointly trains two types of structure, which are x-Rep-x and Rep-x-Rep, to emphasis on the cycle consistency. Thus the bijection can be achieved between the latent and data space.}
	\label{structure}
\end{figure}

Efforts have been put on learning the bidirectional mapping in an adversarial way. InfoGAN \cite{chen2016infogan} is proposed to address the problem of uninformative latent space of GAN, by disentangling the latent variables, and maximizing the mutual information between a subset of the variables and the observations. InfoGAN is able to learn a representation with semantic meaning in a fully unsupervised way. However, faithful reconstruction cannot be achieved by InfoGAN. Another model named Adversarial Autoencoder (AAE)\cite{makhzani2015adversarial} performs variational inference by matching the aggregated posterior distribution with the prior distribution using an adversarial loss. The autoencoder-like structure guarantees a good reconstruction performance, but the generation using the sampled latent variables is not faithful enough. BiGAN\cite{bigan} and ALI\cite{ali} both propose an encoder (inference network) and decoder (generative network), and seek to match the joint distributions of latent variables and data from the two networks. However, the objective functions do not constraint on the relationship between the latent variables and the observations, which results in an unsatisfied reconstruction performance. ALICE \cite{alice} resolves this non-identifiability issues by additionally optimizing the conditional entropy. But it does not learn a disentangled latent space for semantic interpretation and knowledge discovery. 

Bi-directional mapping is also addressed in some applications like image domain transformation or image semantic editing. In BiCycleGAN\cite{bicycle}, the authors differentiated two models cVAE-GAN and cLR-GAN and explained the hybrid model in an intuitive way (regarding to real or fake sampling). It does not encode interpretable information into the latent vector, but directly concatenates the vector with the images from another domain. crVAE \cite{channel} could only demonstrate the semantic meaning of latent vector from visual inspection. IAN\cite{neural_photo_editing} proposed a hybridization of VAE and GAN to solve the semantic photo editing problem by improving the representation capability of latent space without increasing its dimensionality. The decoder of VAE is used as generator of GAN, and hidden layer outputs of discriminator are used to quantify reconstruction loss, which was showed to improve the reconstruction quality. However no cycle consistency was enforced and the latent space was not disentangled. DTN\cite{unsupervised_image_generation} applied a similar structure for image domain transfer, while the latent space is not constrained to a regularized distribution, thus random generation tasks were not performed. 

In this paper, we seek to learn a generic interpretable representation and bidirectional network which is capable of reconstruction, generation and clustering at the same time. A model supporting all these capabilities is important for data analysis and transmission. Reconstruction ability will help data compression while transmitting. Clustering and generation ability will benefit the natural analysis of complicated data without human prior knowledge. 

we first perform a theoretical analysis of two types of structures, which are x-Rep-x and Rep-x-Rep. We identify their advantages and disadvantages respectively by studying the loss functions they try to minimize, and relate it to the mutual information and conditional entropy in the information theory \cite{zhao2017infovae}. Then we propose a novel model involving the concept of cycle consistency \cite{cyclegan,dualgan,discogan} to combine those two structures, which is able to achieve a better overall performance in terms of unsupervised classification accuracy, data reconstruction and generation by learning a useful generic disentangled latent space. Finally, we show and analyze the effectiveness of this new model on the image datasets like MNIST, FashionMNIST, CelebA and SVHN.

\section{Related Work}

In this section, we review the objectives of the two general structures utilized by representation learning, which are x-Rep-x and Rep-x-Rep, as shown in figure \ref{structure}. Here we denote the parameter of the generator (from $z$ to $x$) as $\theta$, and that of the encoder (from $x$ to $z$) as $\phi$. 

\subsection{x-Rep-x Structure}
The x-Rep-x structure is well known as autoencoder, and the most popular instance is Variational Autoencoder (VAE) \cite{vae1,vae2,vaet}. Recall that in the VAE, the variational lower bound it optimizes is,
\begin{equation}
\mathcal {L}(\theta, \phi; x)= KL(q_\phi(z|x)||p_\theta(z)) - \E_{q_\phi(z|x)}[\log{p_\theta(x|z)}]
\end{equation}
The first term is identified as the regularization term which tries to match $q_\phi(z|x)$, the posterior of $z$ conditional on $x$, to a target distribution $p_\theta(z)$ using the KL divergence. The second term represents the reconstruction loss, namely given the data $x$, generating the latent representation $z$, and then using this $z$ to reconstruct the data. Notice that the loss term above is for a specific data point $x$. To get the loss inside a training batch, we need to average it over $x$, namely
\begin{equation}
\begin{aligned}
\mathcal {L}_{VAE} = & \E_{p_{data}(x)}[\mathcal {L}(\theta, \phi; x)]\\
 = &  \E_{p_{data}(x)}[KL(q_\phi(z|x)||p_\theta(z))]\\
 & - \E_{p_{data}(x)}[\E_{q_\phi(z|x)}[\log{p_\theta(x|z)}]]
\end{aligned}\label{loss_aae}
\end{equation}

Adversarial Autoencoder\cite{makhzani2015adversarial} is a related work of the VAE. The loss function of AAE is similar to VAE, except for the regularization term $KL(q_\phi(z|x)||p_\theta(z))$ is replaced by an adversarial learning process (represented by JS Divergence) on the aggregated posterior distribution. Therefore, the objective function for AAE becomes,
\begin{equation}
\begin{aligned}
\mathcal{L}_{AAE} &= JS(q_{\phi}(z)||p_{\theta}(z)) - \E_{q_{\phi}(z,x)}[\log p_{\theta}(x|z)]
\end{aligned} \label{eq:aae_loss}
\end{equation}

InfoVAE\cite{zhao2017infovae} generalizes the regularization term of AAE to a divergence family, and justify the richer information it provides in the latent code sampled from the aggregated posterior distribution. The author also proved that the latent space learned by InfoVAE (or the variants AAE) does not suffer from exploding problem and uninformative latent modeling. However, unsupervised generation with disentangled latent vector was not reported in the original VAE, AAE or InfoVAE paper.

\subsection{Rep-x-Rep Structure}
The Rep-x-Rep structure is derived from the vanilla GAN \cite{goodfellow2016nips} , by making the discriminator output not only the real or fake label, but also the semantic  representation \cite{chen2016infogan,acgan,featurelearninggan,catgan}.  

One example of this structure utilized for unsupervised learning is InfoGAN (figure \ref{structure}), which basically adds an information maximizing term on top of the vanilla GAN \cite{goodfellow2016nips} so that the generator is forced to use all the information contained in the input when generating sample data points. In the original InfoGAN paper, the latent vector is disentangled into categorical, continuous and noise parts, and the discriminator will output the categorical and continuous parts to achieve the mutual information maximization. 
 The loss function of such Rep-x-Rep structure can be written as,
 
\begin{equation}
\begin{aligned}
	L_{InfoGAN} =& JS(p_{\theta}(x)||p_{data}(x))\\
    & -\E_{p_\theta(z)}[{\E_{p_\theta(x|z)}[\log{q_\phi(z|x)}}]
\label{loss_infogan}
\end{aligned}
\end{equation}
 
InfoGAN achieves a fully unsupervised representation learning with disentangled semantic latent space. Both the generation and clustering performance is impressive, but the reconstruction quality of input images was not satisfied to report. 
\begin{figure}[t]
\begin{center}
    \includegraphics[width=1\linewidth]{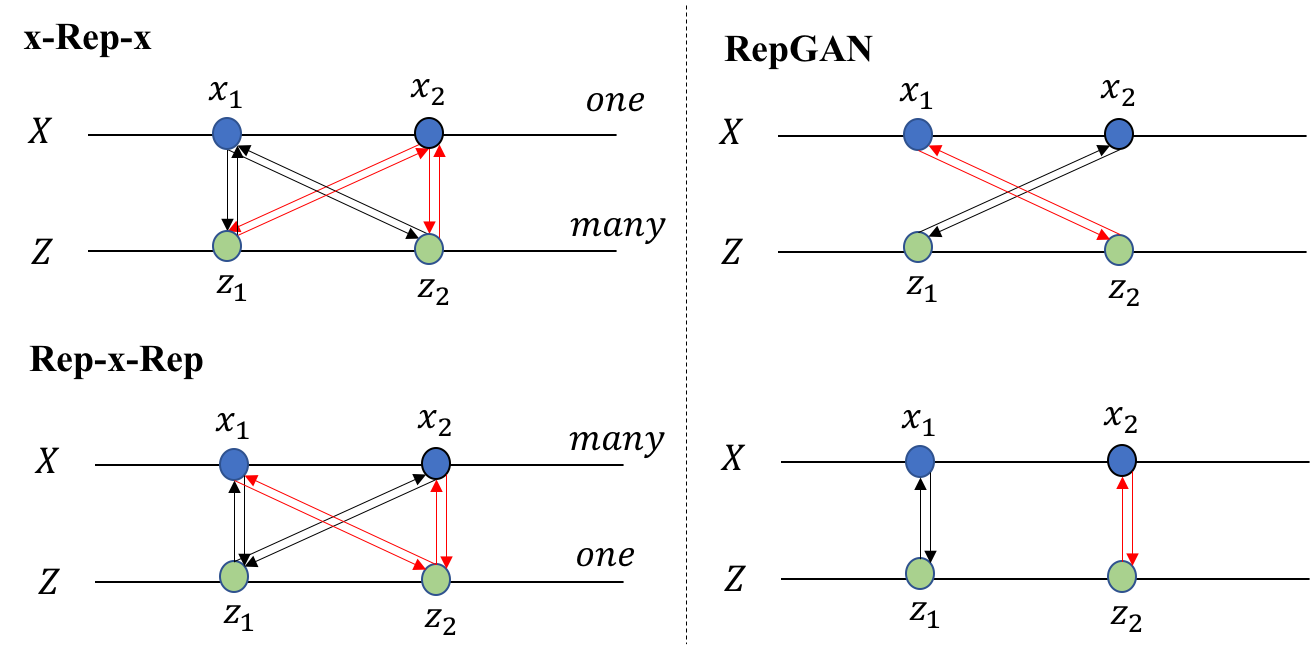}
\end{center}
\caption{Mapping relations of x-Rep-x ,Rep-x-Rep and proposed RepGAN. Cycle consistency is involved for a deterministic bijection better for both reconstruction and generation.}
\label{mapping}
\end{figure}

\section{RepGAN}
In this section, we derive the correlation between the loss function of previous models and conditional entropy, and compare their influence on the learned representation. Then we integrate the loss and propose a novel model called Representation GAN to learn a useful disentangled latent space.
\subsection{Conditional Entropy}\label{sec:ce}
The conditional entropy measures the uncertainty of one random variable given the other. For example, $H(X|Z)$ quantifies the uncertainty of the observation space $X$ given the latent space $Z$, which can be formulated as,
\begin{equation}
\begin{aligned}
H(X|Z) = &- \E_{p_{data}(x)}[\E_{q_\phi(z|x)}[\log{q_\phi(x|z)}]]\\
=& - \E_{p_{data}(x)}[\E_{q_\phi(z|x)}[\log{p_\theta(x|z)}]]\\
&- \E_{p(z)}[KL(q_\phi(x|z)||p_\theta(x|z))]\\
\leq & - \E_{p_{data}(x)}[\E_{q_\phi(z|x)}[\log{p_\theta(x|z)}]]
\end{aligned}
\end{equation}

Comparing the upper bound of conditional entropy with the second term in VAE/AAE loss (equation \ref{loss_aae}), we can see that they are exactly the same expression. As a result, the upper bound of the conditional entropy $H(X|Z)$ is equivalent to the reconstruction term of VAE/AAE objective function. Minimizing the upper bound of the conditional entropy also turns out to be maximizing the mutual information $I(X,Z)=H(X)-H(X|Z)$, if the entropy of the data distribution $H(X)$ is assumed to be fixed.

Again the formulation of conditional entropy $H(Z|X)$ is,
\begin{equation}
\begin{aligned}
H(Z|X) = &- \E_{p_{\theta}(z)}[\E_{p_\theta(x|z)}[\log{p_\theta(z|x)}]]\\
=& - \E_{p_{\theta}(z)}[\E_{p_\theta(x|z)}[\log{q_\phi(z|x)}]]\\
&- \E_{p_{data}(x)}[KL(p_\theta(z|x)||q_\phi(z|x))]\\
\leq & - \E_{p_{\theta}(z)}[\E_{p_\theta(x|z)}[\log{q_\phi(z|x)}]]
\end{aligned}
\end{equation}

We see that the reconstruction loss term in equation \ref{loss_infogan} is exactly the same as the upper bound of the conditional entropy $H(Z|X)$. This loss function tries to minimize the conditional entropy $H(Z|X)$ and consequently maximizing the mutual information $I(X,Z) =H(Z)-H(Z|X)$.

\subsection{Comparing x-Rep-x and Rep-x-Rep}
Now we study the property of the loss function of x-Rep-x and Rep-x-Rep. Both the two structures try to maximize the mutual information $I(X,Z)$ by minimizing conditional entropies. However since conditional entropy is not symmetric, those two structures show different focuses.x-Rep-x structure minimizes the conditional entropy $H(X|Z)$. It is trained to decrease the uncertainty of $x$ given $z$. As shown in figure \ref{mapping}, it demonstrates a stochastic mapping from $x$ to the latent representation $z$. Comparing equation \ref{loss_infogan} to equation \ref{loss_aae}, we can see that they are symmetric to each other. As a result, we can use the same argument as in the x-Rep-x case to conclude that Rep-x-Rep actually maps multiple data points $x$ back to the same latent representation $z$ as shown in figure \ref{mapping}. 

In conclusion, we show that x-Rep-x maps multiple points in latent space to a single point in data space, whereas Rep-x-Rep maps multiple points in data space to a single point in latent space. Therefore, x-Rep-x is good at reconstruction (when the latent space is large enough). The classification performance of it is not guaranteed though. On the other hand, x-Rep-x is good at classification, because different digits with subtle differences can be put into the same category, which makes the classifier robust to noises and small style changes. But the reconstruction performance is not guaranteed. 

\begin{figure*}[ht!]
	\begin{center}
		\includegraphics[width=1\linewidth]{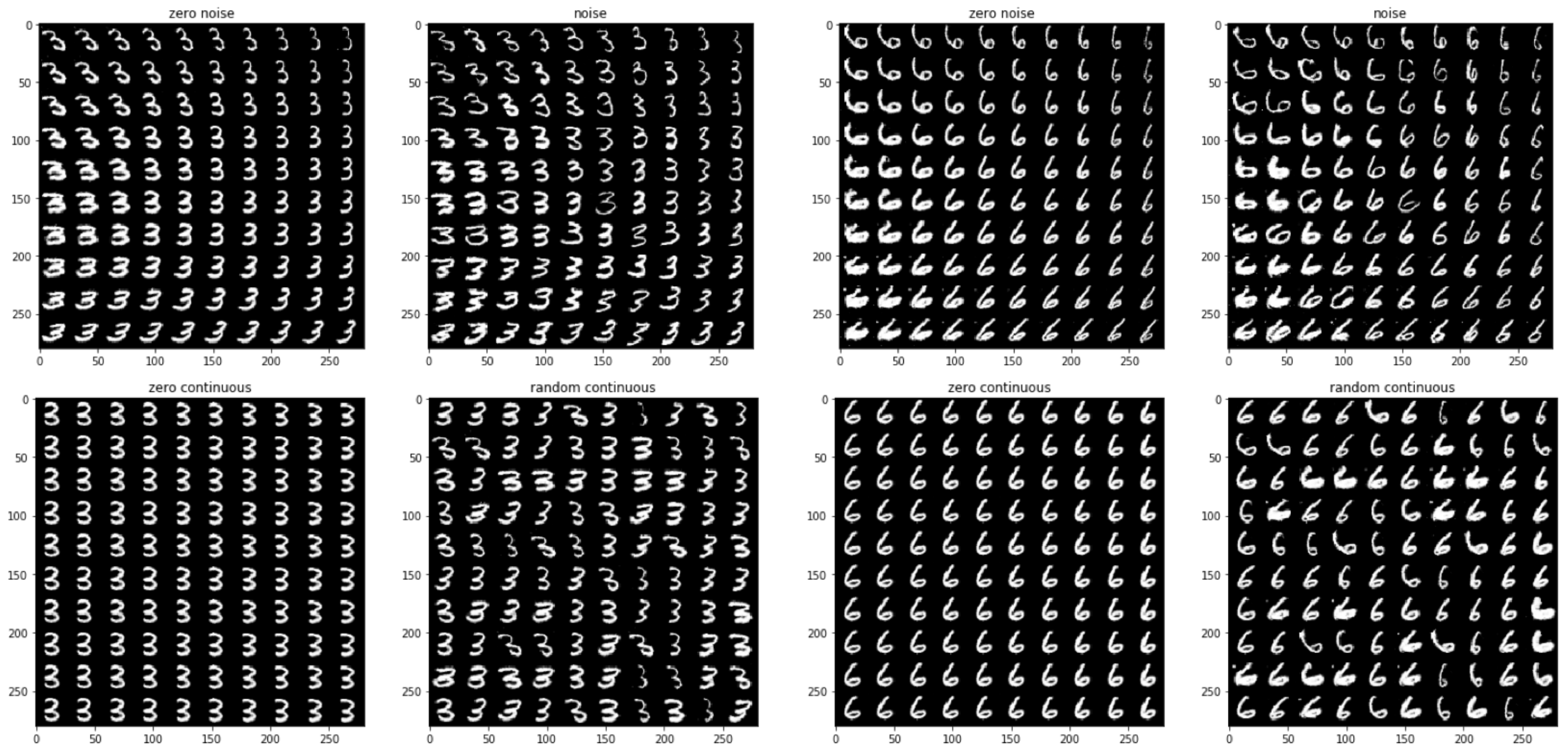}
	\end{center}
	\caption{First row: traversing on continuous variable $s$ with zero noise, and then controlling $s$ and adding noise $n$. Using the same noise batch for different clusters, but it shows different variants. Thus noise variable is cluster-dependent. Second row: traversing on the first two dimensions of noise vector with zero $s$. Tiny changes of samples can be visualized. Then we add same random continuous variable value for distinct clusters, and identical changes are illustrated. $s$ is cluster-independent and corresponding to commonly shared attributes: slant and thickness.}
	\label{generation_mnist}
\end{figure*}

\begin{figure*}[ht!]
	\begin{center}
		\includegraphics[width=1\linewidth]{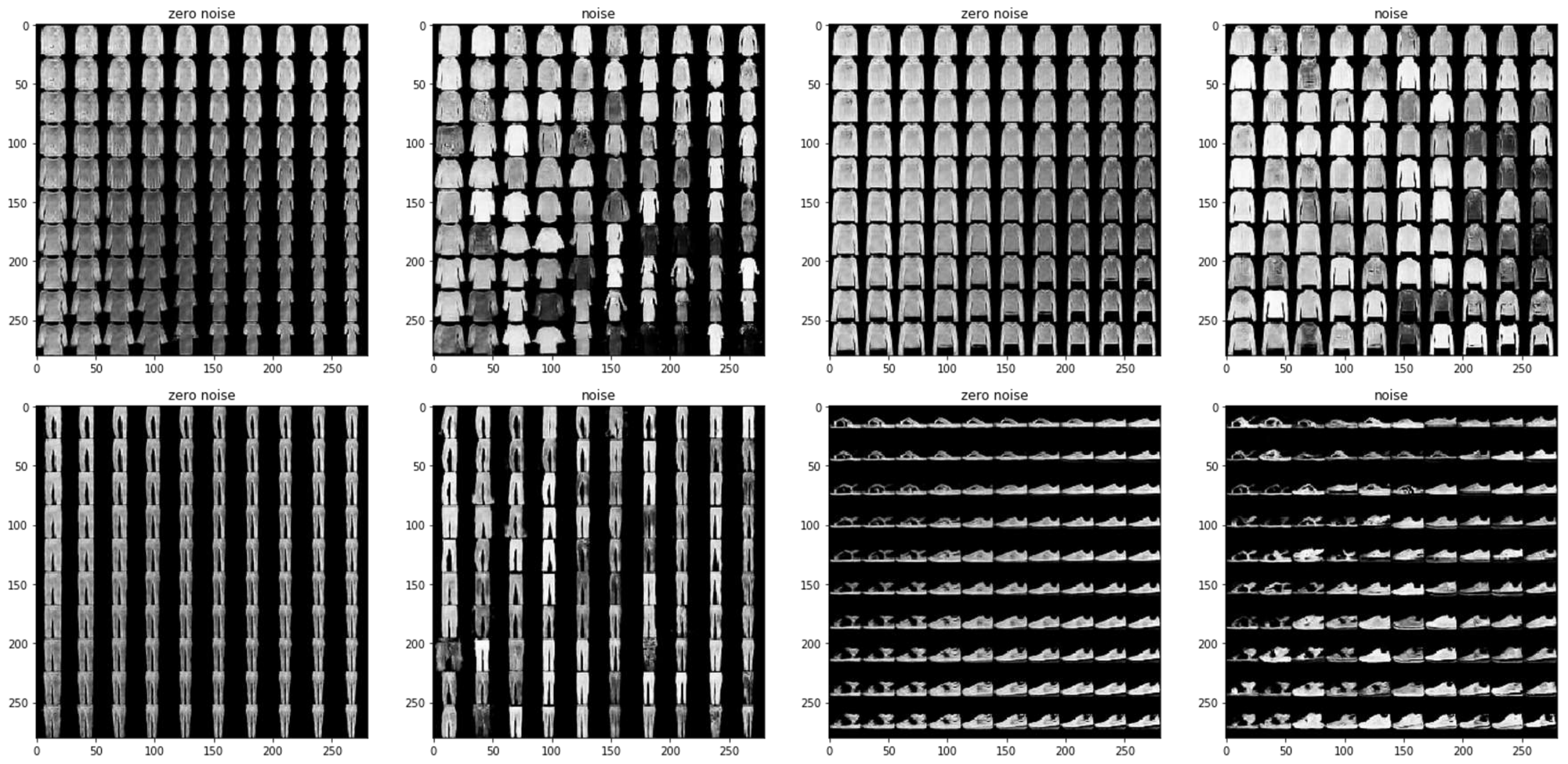}
	\end{center}
	\caption{Generated samples on FashionMNIST dataset. Similar to MNIST, we can see a smooth style transition when noise is set to zero. Compared to MNIST dataset, noise variables have a larger impact on generated images on fashionMNIST. In addition, since the RepGAN on fashionMNIST is trained unsupervised, different categories with similar appearance may get classified as same category, as shown on the bottom right sub-figure where sandals and sneakers are confused.}
	\label{generation_fmnist}
\end{figure*}

Actually, if we follow the design of the latent vector in the original InfoGAN paper, the reconstruction of InfoGAN cannot be good because the noise at the input of InfoGAN is not present at the output, which means subtle information describing the details of the image is discarded during reconstruction. In our experiments, we find out that the noise actually changes the generated image greatly, as shown in figure \ref{generation_mnist} and \ref{generation_fmnist}. If the noise is simply discarded, the reconstructed images will be almost not the same.

To further understand the mapping relation of x-Rep-x and Rep-x-Rep, suppose in the discrete case, we notice that the second term in equation \ref{loss_aae} is minimized to zero when $p_\theta(x|z)$ equals one for all $z$, whereas those $z$ are actually the output from the encoder of the AAE. Thus in the optimal case, the probability mass function $q_\phi(z|x)$ should have disjoint support for different given $x$ \cite{zhao2017infovae}, but it is not optimized to 1 for a specific $z$. That is the reason why one data point $x$ can be mapped to different $z$, and multiple $z$ can be used to reconstruct the same $x$. Similar explanations apply to the mapping property of Rep-x-Rep.

\begin{figure}[t]
	\begin{center}
		\includegraphics[width=1\linewidth]{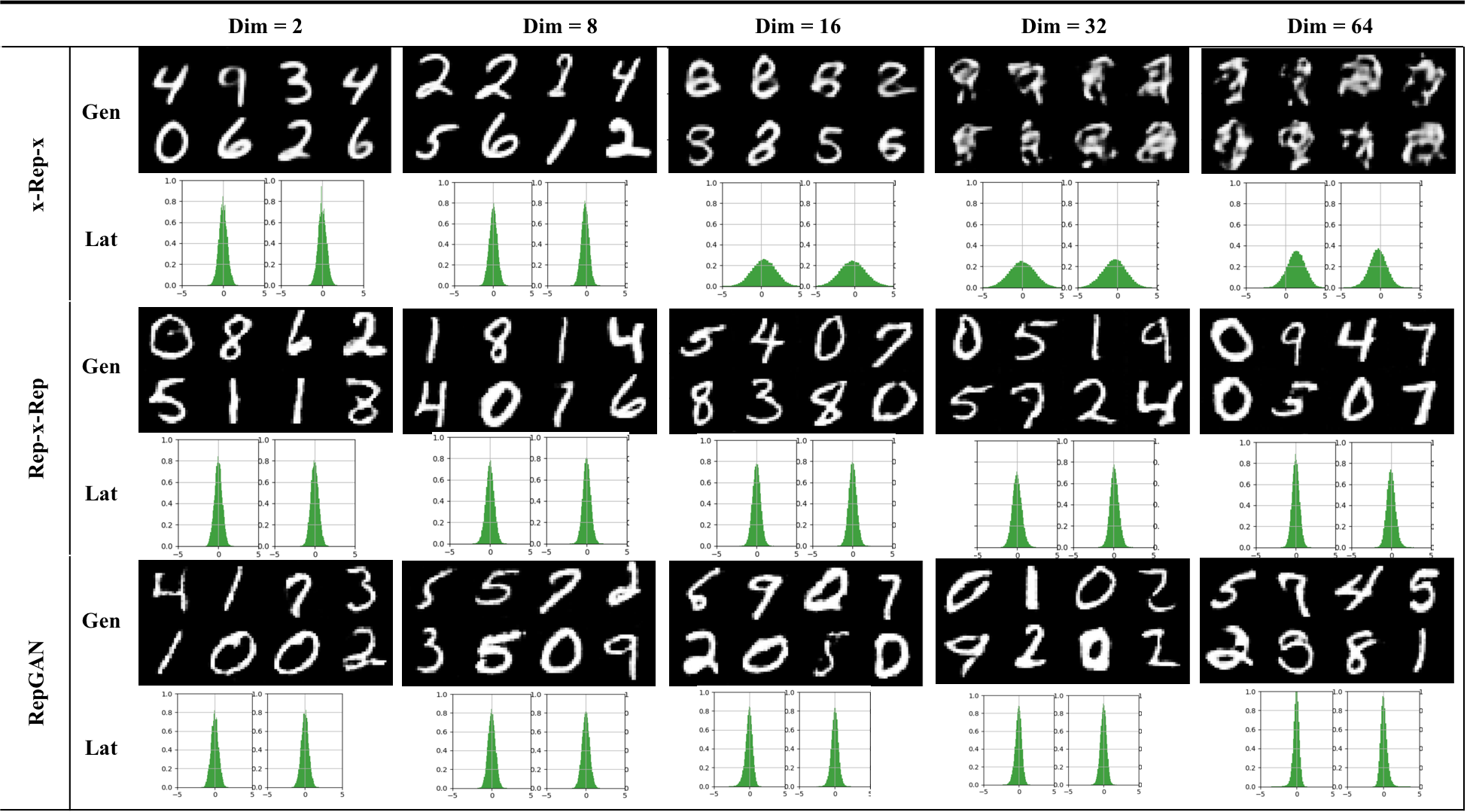}
	\end{center}
	\caption{Generated samples and the randomly selected learned latent distribution of all the three models.The target latent space is zero-mean Gaussian with 0.5 variance.}
	\label{fig:gen}
\end{figure}

\begin{figure}[t]
	\begin{center}
		\includegraphics[width=1\linewidth]{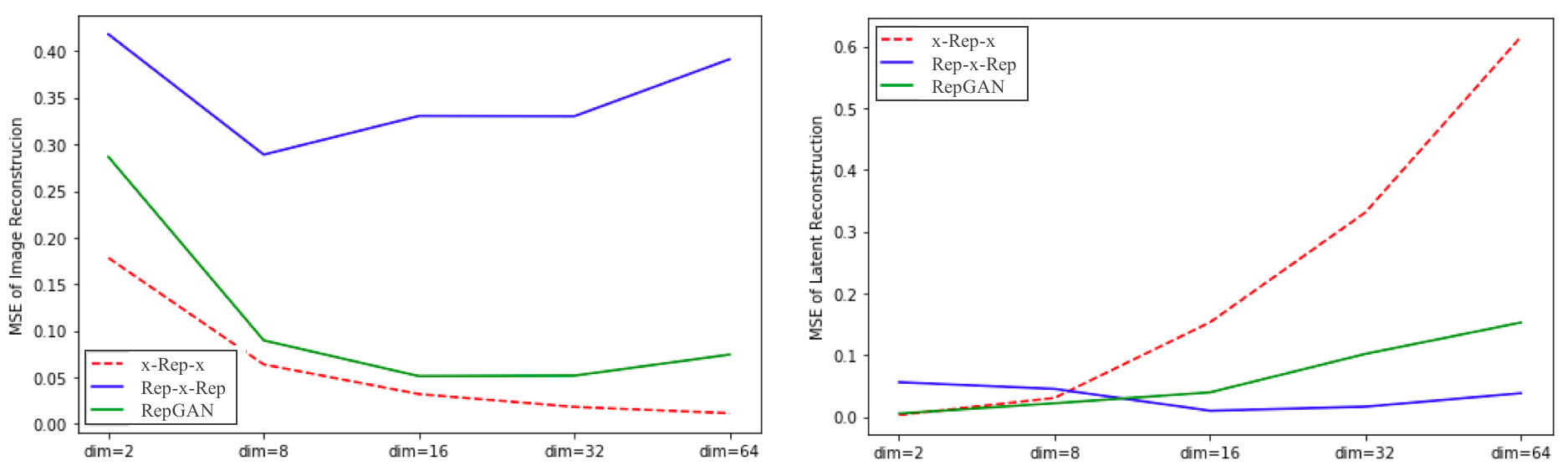}
	\end{center}
	\caption{Reconstruction error and latent vector error curve for the three models across the latent dimensions. Left: Data reconstruction error in terms of MSE. Right: Illustrating the ability of generation by showing latent Reconstruction error in MSE. }
	\label{fig:curve}
\end{figure}

Another problem is the dimension of the latent vector. Lower latent dimension will suffer from insufficient representation ability, while higher latent dimension will increase the difficulty of distribution regularization using adversarial learning or KL divergence. The drawbacks are shown in figure \ref{fig:gen} and \ref{fig:curve}.

\subsection{Model Structure}
In order to combine the strength of both two structures, we propose to train x-Rep-x and Rep-x-Rep together with shared parameters elegantly, so that the new model can achieve good classification and reconstruction performance at the same time. The network architecture is illustrated in figure \ref{structure}. Specifically, the encoder (x2Rep) of x-Rep- and Rep-x-Rep are the same module sharing parameters. Likewise, the decoder (Rep2X) of the two structures are also the same module sharing parameters. During the training, we train the model alternatively between x-Rep-x-fashion and Rep-x-Rep-fashion so that the classification accuracy can be improved by Rep-x-Rep training while reconstruction performance can be improved by x-Rep-x training. The training of Rep-x-Rep is emphasized with five times iterations, which experimentally gives better result. The latent vector is split into three subsets, a categorical variable $c$, a continuous variable $s$ and a noise $n$. Continuous and noise variable could be sampled from Gaussian distribution. The full objective function for RepGAN is,\\ 
\begin{equation}
\begin{aligned}
\mathcal{L}_{RepGAN} = &  JS(q_{\phi}(c)||p(c))\\
		 & +JS(q_{\phi}(s)||p(s))\\
		 & +JS(q_{\phi}(n)||p(n))\\ 
         & -\E_{q_{\phi}(z,x)}(\log p_{\theta}(x|z))\\
	     & +JS(p_{\theta}(x)||p_{data}(x))\\
         & -\E_{p(c)}[\E_{p_\theta(x|c)}(\log{q_\phi(c|x)})]\\
         & -\E_{p(s)}[\E_{p_\theta(x|s)}(\log{q_\phi(s|x)})]
\end{aligned}
\end{equation}

In practical implementation, we rewrite the objective function as,
\begin{equation}
\begin{aligned}
\mathcal{L}_{RepGAN} = & \mathcal{L}_{AdvC} + \mathcal{L}_{AdvS} + \mathcal{L}_{AdvN} +
\mathcal{L}_{AdvX}\\
& + \mathcal{L}_{RecX} +
\mathcal{L}_{RecC} +
\mathcal{L}_{RecS}
\end{aligned}
\end{equation}

where $\mathcal{L}_{RecX}$ is computed as $L_2$ norm for image reconstruction, $\mathcal{L}_{RecC}$ represents cross-entropy loss, and $\mathcal{L}_{RecS}$ is negative log-likelihood for Gaussian loss with re-parameterization tricks as InfoGAN. The model structure for experiment is summarized in table \ref{network_structure}.The stride for each convolution layer is always 2, and we refer the structure design and optimization tricks as WGAN\cite{preWGAN,wgan}. For learning rate, we used $5e\text{-}4$, $1e\text{-}3$, $2e\text{-}4$ for the generators, x-Rep-x discriminator and Rep-x-Rep discriminator on MNIST dataset. For fashionMNIST, we used $5e\text{-}5$, $1e\text{-}3$, $2e\text{-}5$. For SVHN, we used $1e\text{-}4$, $1e\text{-}3$, $2e\text{-}5$.

\begin{table}[t]
\caption{Network Structure of RepGAN}
\label{network_structure}
\begin{center}
\begin{tabular}{llll}
\multicolumn{1}{c}{\bf encoder}  
&\multicolumn{1}{c}{\bf decoder} 
\\ 
In 28x28x1 &In 32x1\\
4x4x64 conv, LReLU,BN & FC1024 ReLU,BN  \\
4x4x128 conv,LReLU, BN& FC7x7x128 ReLU,BN\\
FC 1024 LReLU, BN & 4x4x64 deConv,ReLU,BN \\
c: FC 10 softmax, BN & 4x4x1 deConv, Sigmoid\\
s mean:  FC 2 LRelu, BN \\
s sigma: FC 2 LRelu, BN, exp()\\
n:     FC 20 LRelu, BN\\

\multicolumn{1}{c}{\bf Dz}  
&\multicolumn{1}{c}{\bf Dx} 
\\ 
In c/s/n & In 28x28x1 \\
FC3000 LReLU & 4x4x64 conv, LReLU\\
FC3000 LReLU & 4x4x128 conv, LReLU, BN \\
FC1 raw output (WGAN) &FC1024 LRelu, BN \\
&FC 1 sigmoid\\

\end{tabular}
\end{center}
\end{table}

\begin{figure}[t]
	\begin{center}
		\includegraphics[width=1\linewidth]{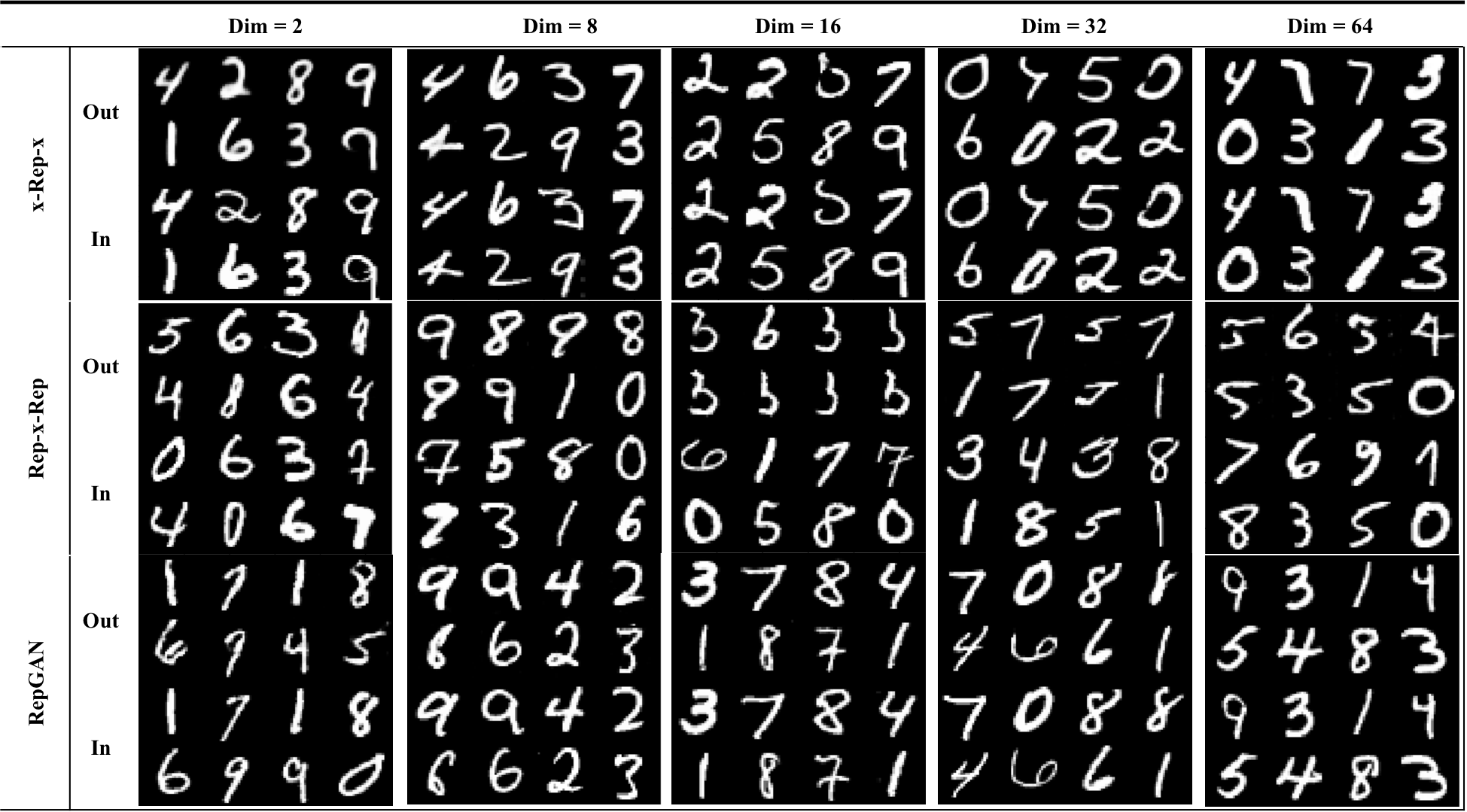}
	\end{center}
	\caption{Reconstruction visualization of x-Rep-x, Rep-x-Rep and RepGAN with Gaussian latent space. x-Rep-x and RepGAN achieves identically good reconstruction, but Rep-x-Rep cannot recover the original input image good enough. The reconstructed images are blur when the latent dimension of x-Rep-x is low, but better with RepGAN.}
	\label{fig:rec}
\end{figure}

\begin{table}[t]
\caption{Testing Accuracy for Unsupervised Classification}
\label{tab:acc}
\begin{center}
\begin{tabular}{lllll}
\multicolumn{1}{c}{\bf Model}  &\multicolumn{1}{c}{\bf MNIST} &\multicolumn{1}{c}{\bf MNIST} &\multicolumn{1}{c}{\bf FMNIST} &\multicolumn{1}{c}{\bf FMNIST} \\
&(Acc) & (MSE) & (Acc) & (MSE) \\
\\ \hline \\
x-Rep-x         &86.92\%   &\bf 0.007   &57.30\% &0.015\\
Rep-x-Rep            &95\%   &0.07  & 53.81\% &0.098\\
VADE          &94.46\%   &None & None &None\\
DEC         &84.30\%   &None & None &None\\
RepGAN            &\bf 96.74\%   &0.02 & \bf 58.64\% & \bf 0.013\\

\end{tabular}
\end{center}
\end{table}
\begin{figure}[ht]
	\begin{center}
		\includegraphics[width=1\linewidth]{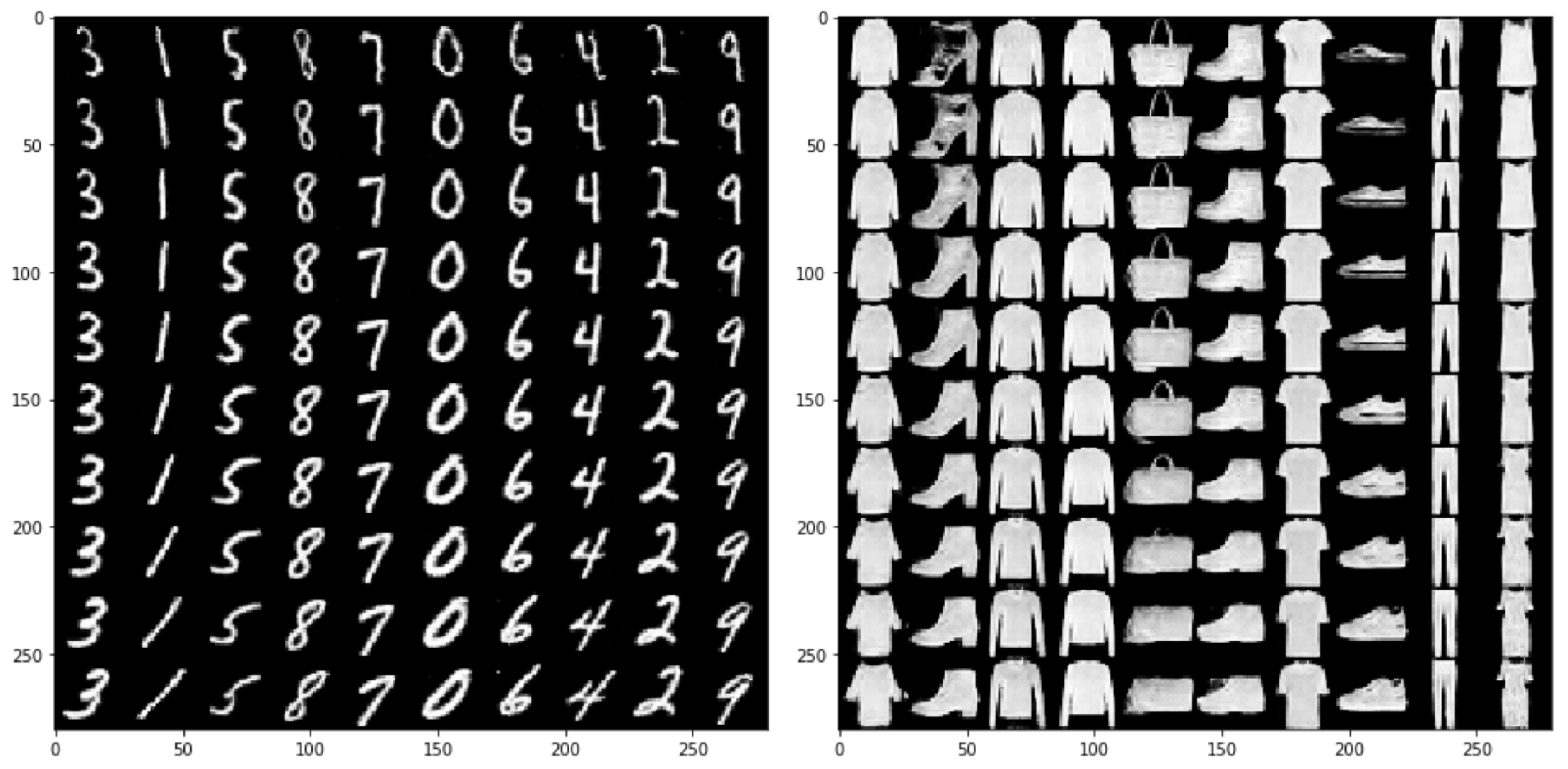}
	\end{center}
	\caption{Varying the categorical variable, the RepGAN can cluster different types of images in MNIST and FashionMNIST in a fully unsupervised way. The categorical variable varies along the column, and the continuous variable varies along the row.}
	\label{fig:cat}
\end{figure}

\begin{figure}[ht]
	\begin{center}
		\includegraphics[width=1\linewidth]{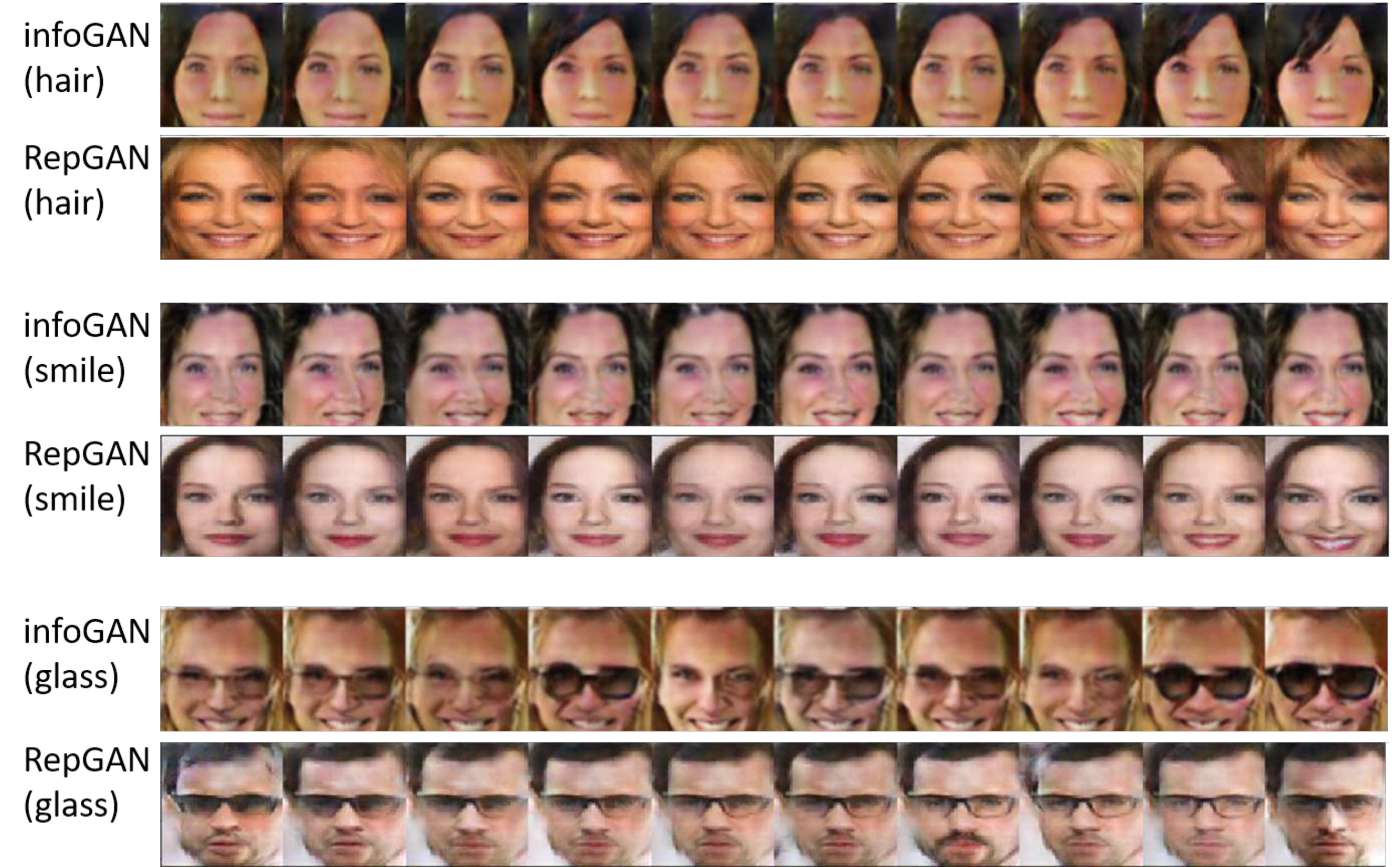}
	\end{center}
	\caption{Comparison of different attributes generation ability of RepGAN and InfoGAN. RepGAN generates better-quality images than infoGAN.}
	\label{fig:celebA}
\end{figure}

\begin{figure}[ht]
	\begin{center}
		\includegraphics[width=1\linewidth]{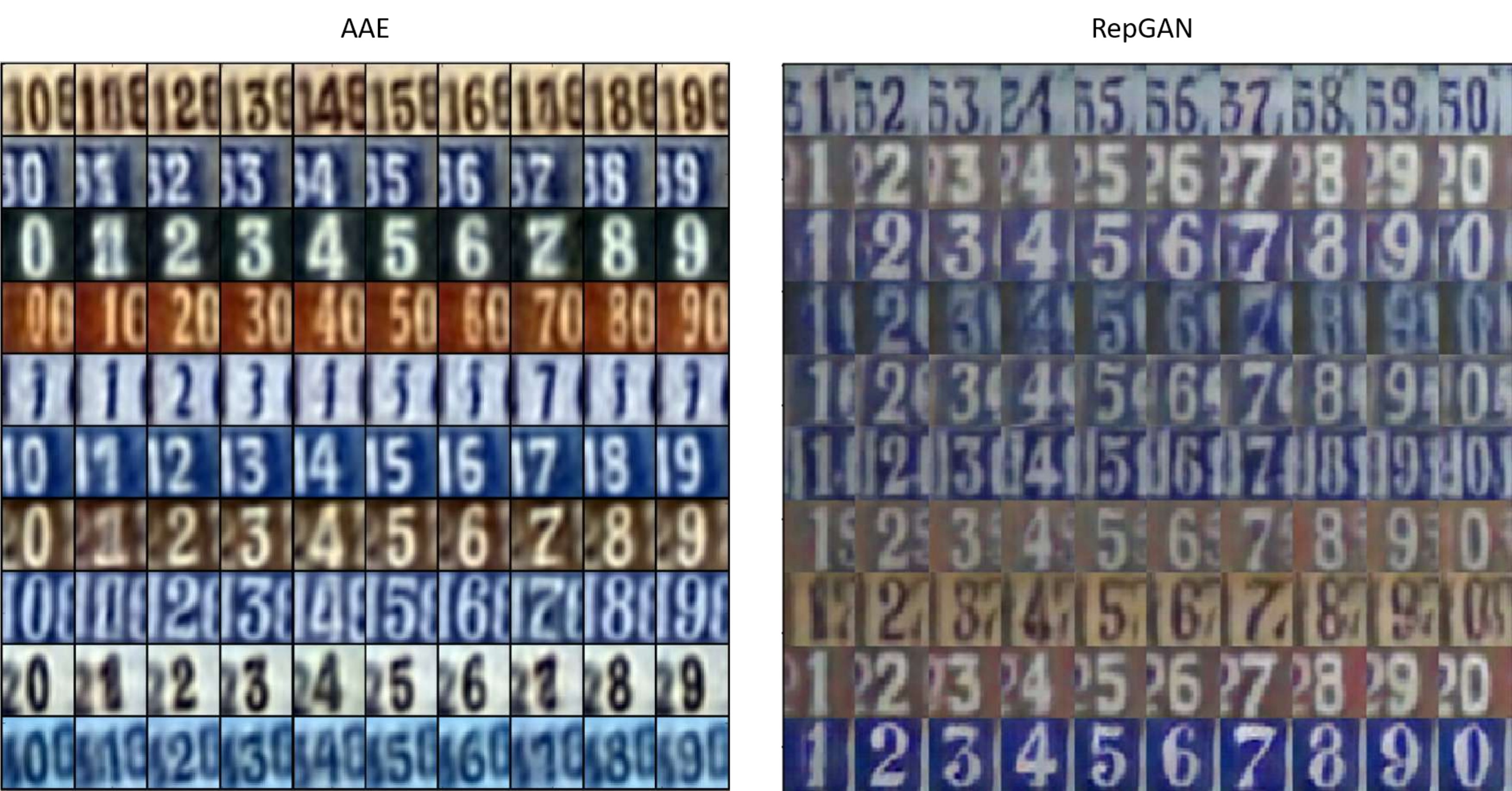}
	\end{center}
	\caption{Comparison of generated SVHN samples between x-Rep-x (AAE) and RepGAN. RepGAN generates sharper images than x-Rep-x trained structure.}
	\label{fig:compare_SVHN}
\end{figure}

\begin{figure*}[t]
	\begin{center}
		\includegraphics[width=1\linewidth]{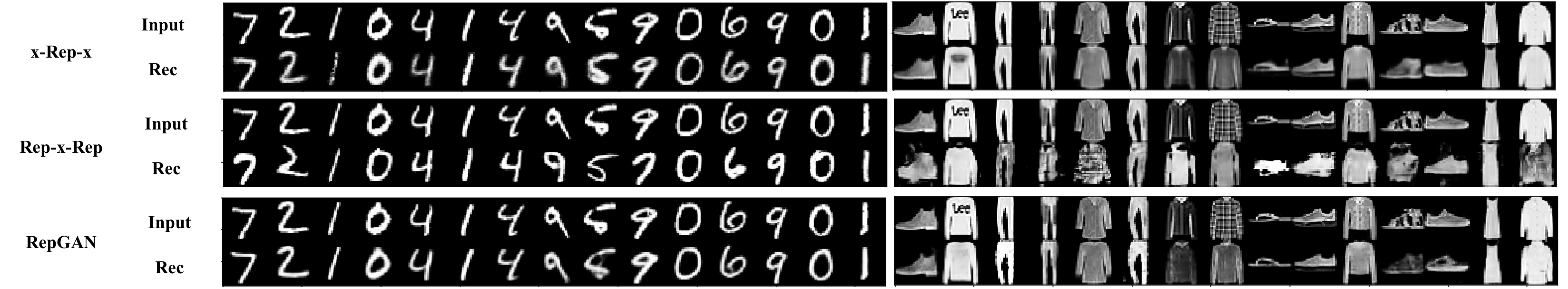}
	\end{center}
	\caption{Reconstruction performance of x-Rep-x, Rep-x-Rep and RepGAN with disentangled latent space. The first row of each group of images are the input, and the second row is the reconstruction. Rep-x-Rep cannot achieve a faithful reconstruction, and x-Rep-x only recovers blurred images if the latent dimension is not big enough. RepGAN achieves sharper and more clear reconstruction. }
	\label{fig:recon}
\end{figure*}

\section{Experiments}
We tested the three models x-Rep-x, Rep-x-Rep and RepGAN on MNIST\cite{mnist}, FashionMNIST\cite{fashionmnist}, and SVHN\cite{svhn} dataset. We conducted two sets of experiments with different types of the latent variable $z$. First, $z$ is not disentangled but directly sampled from an isotropic Gaussian distribution. In this experiment, we investigate the theory of mapping discussed in last section. Second, like the original InfoGAN, we split the latent vector $z$ into three slots: a one-hot vector $c$ sampled from a categorical distribution, a continuous vector $s$ sampled from Gaussian, and a random noise $n$. In additional to reconstruction and generation performance, we demonstrate the unsupervised clustering performance of RepGAN, and noise importance. 

\subsection{Gaussian Latent space}
We first implement the x-Rep-x, Rep-x-Rep, and RepGAN with a single entangled latent space using a latent vector sampled from isotropic Gaussian distribution with zero mean and 0.5 variance. We vary the dimension of the latent vector to 2, 8, 16, 32 and 64, and compare the reconstruction performance of image or latent space of all the three models. Training and testing is conducted on MNIST dataset.

\subsubsection{Image Reconstruction}

The image reconstruction result is computed by organizing the structure like x-Rep-x after training each model, and feeding the real data sample to the input. The visualization is shown in figure \ref{fig:rec}. x-Rep-x achieves a better reconstruction ability than Rep-x-Rep, and RepGAN is almost as good as x-Rep-x. As shown in the figure \ref{fig:rec} and \ref{fig:curve}, Rep-x-Rep has a bad ability of reconstruction, and for all the latent dimensions, the error keeps the highest among the three models. That is because the loss definition of Rep-x-Rep does not put constrains on the image reconstruction.

\subsubsection{Latent Reconstruction and Generation}
For latent reconstruction evaluation, we follow the structure of Rep-x-Rep for testing. After training all the models, we reorganize the network structure, and feed a sampled latent vector $z$ into the network. we plotted the MSE of the latent code to examine the ability of latent regularization and the exist of mode collapse of the models in figure \ref{fig:curve}. For x-Rep-x trained one, the error becomes large when the latent dimension is high because of (1) heavy mode collapse: given different z, the model generates identical x. This is illustrated in the objective, and (2) unsatisfied latent regularizing like VAE. In this case, when we sample from a true prior distribution, the x-Rep-x model cannot generate good-quality images. That is because the high-quality manifold shifted.As shown in figure \ref{fig:gen}, autoencoder cannot generate high-quality samples when the latent dimension is too large, and cannot generate sharp samples when the dimension is small. The model fails to learn a good latent distribution when the latent dimension is larger or equal to 16. 

However, Rep-x-Rep trained structure and RepGAN achieve an identical good performance for latent space modeling and new sample generation. The generated images are also sharp and clear. All the images in figure \ref{fig:gen} are randomly sampled from the generation results. Compared with x-Rep-x and Rep-x-Rep which can only guarantee either recognition and generation, the proposed RepGAN can simultaneously achieve the two capabilities by constraining on two conditional entropy, and the mapping between the latent variable and real data shrinks to a bijection.  

\subsection{Disentangled Latent space}
In this experiment, we disentangle the latent space and follow the original structure of InfoGAN. $c$, $s$ and $n$ have dimension 10, 2, and 20 respectively. In addition to reconstruction and generation, we compare the unsupervised clustering performance. We also investigate the importance of the noise. 

\subsubsection{Unsupervised Learning}

When evaluating the unsupervised clustering accuracy, we set the continuous and noise vector to zero, and generate the cluster head of each clusters. Then we searched in the training set to find the closest sample with the cluster head, and assigned the label of that sample to the whole cluster. Finally, we computed the accuracy based on the assigned cluster labels. Table \ref{tab:acc} shows the classification accuracies of comparable models like VADE\cite{vade} and DEC\cite{dec} on MNIST and FashionMNIST. The Rep-x-Rep and RepGAN are able to achieve an average accuracy of 95\% or 96\%, which is much higher than the x-Rep-x, which only achieves 87\%. For FashionMNIST, the classification accuracy is low due to the high similarity of images assigned by different category labels. This experiment result is consistent with our theoretical analysis, which is Rep-x-Rep is better for classification than x-Rep-x. RepGAN, being the elegant combination of the two structures, successfully preserved the ability of Rep-x-Rep for clustering and generation, and x-Rep-x for reconstruction. 

The qualitative evaluation of reconstruction and generation ability of RepGAN is shown in figure \ref{fig:cat} and \ref{fig:recon}. By fixing the categorical code, the model is able to generate any samples belonging to this cluster. And by changing the continuous value, the model learns the manifold of the styles. While reconstructing, RepGAN achieves a more faithful reconstruction than Rep-x-Rep, and sharper images than x-Rep-x. 

We also compare our generated image on CelebA with infoGAN in figure \ref{fig:celebA}. By using the same latent space configuration as infoGAN, namely 10 categorical variables where each one is 10-dim OneHot vector, we are able to achieve a better image quality while showing attribute change at the same time. In addition, we compare the generation quality on SVHN between x-Rep-x and RepGAN in figure \ref{fig:compare_SVHN} and shows that RepGAN generates sharper images than an autoencoder. In summary, RepGAN currently does well in all the three tasks: reconstruction, generation, and unsupervised clustering.

\subsubsection{Effectiveness of Noise }

The noise variable is interpreted as representation incompressible information in InfoGAN. We tunnel the noise for intact and plausible image reconstruction during training the x-Rep-x part, since categorical and continuous variable may not be expressive enough for intact reconstruction. The difference between continuous and noise variable is that: the lower-dimensional continuous variable is used to encode the most salient attributes (or largest data variance direction) commonly shared by all the samples (it is enhanced by $L_{RecS}$), while noise is used to encode incompressible or entangled information (it is enhanced by $L_{RecX}$). 

In figure \ref{generation_mnist} and \ref{generation_fmnist}, we demonstrate the effect of continuous v.s. noise variable on generated samples. Specifically, on the first row, we interpolate on the continuous code, and set the noise variable to zero. While varying the continuous variable, the style changes explicitly and smoothly. After adding random noise, in addition to uniformly changed style, more variants are generated. 

On the second row, we interpolate the first two dimension of the noise code, and set the continuous variable to zero. We can see tiny changes of the generated images when traversing on the first two dimensions of noise, and the changes are slightly different for distinct clusters. It demonstrated the information encoded in noise is actually cluster-dependent. If we randomly sample the continuous variable and keep it the same for all the clusters, we can visualize identical changes of the image attributes across clusters (slant and thickness degree). It demonstrated the information encoded in $s$ is actually cluster-independent or cluster-shared. 

\section{Conclusion and Discussion}
In this paper, we analyzed the advantage and disadvantage of two unsupervised machine learning strictures: x-Rep-x and Rep-x-Rep. We showed both theoretically and experimentally that Rep-x-Rep is able to achieve a higher classification accuracy, whereas x-Rep-x is able to get a better reconstruction quality. After that, we combined those structures elegantly in an attempt to take their advantages. We showed on MNIST, FashionMNIST and SVHN dataset that the new model, named RepGAN, is able to achieve both a high classification accuracy and a good reconstruction quality in both the original input space and the latent space. By performing well in both classification and reconstruction, RepGAN is able to learn a good bidirectional mapping between the input space and the latent space, which is a desired property of unsupervised representation learning model. It will be inspiring if it can be utilized for arbitrarily complicated data discovery with more complicated network structures and larger latent dimension, which is left for future work.

\bibliographystyle{ieeetr}
\bibliography{main}

\end{document}